\documentclass[11pt,a4paper]{article}
\usepackage[hyperref]{emnlp-ijcnlp-2019}
\usepackage{times}
\usepackage{latexsym}
\usepackage{url}
\usepackage{velthuis}
\usepackage{multirow}
\usepackage{graphicx}
\usepackage{subfig}

\aclfinalcopy % Uncomment this line for the final submission

\newcommand{\oyeshds}{OurNepali }
\title{Named Entity Recognition for Nepali Language}

\author{Oyesh Mann Singh, Ankur Padia and Anupam Joshi \\
        University of Maryland, Baltimore County \\
        Baltimore, MD, USA \\
        {\tt{\{osingh1, pankur1, joshi\}}@umbc.edu}}

\begin{document}
\maketitle
\begin{abstract}
Named Entity Recognition have been studied for different languages like English, German, Spanish and many others but no study have focused on Nepali language. In this paper we propose a neural based Nepali NER using latest state-of-the-art architecture based on grapheme-level which doesn't require any hand-crafted features and no data pre-processing. Our novel neural based model gained relative improvement of 33\% to 50\% compared to feature based SVM model and up to 10\% improvement over state-of-the-art neural based models developed for languages beside Nepali.
\end{abstract}

\section{Introduction}
Named Entity Recognition (NER) is a foremost NLP task to label each atomic elements of a sentence into specific categories like "PERSON", "LOCATION", "ORGANIZATION" and others\cite{DBLP:journals/corr/abs-1103-0398}. There has been an extensive NER research on English, German, Dutch and Spanish language \cite{DBLP:journals/corr/LampleBSKD16}, \cite{DBLP:journals/corr/MaH16}, \cite{DBLP:journals/corr/abs-1810-04805},
\cite{DBLP:journals/corr/abs-1802-05365},
\cite{akbik-etal-2018-contextual}, and notable research on low resource South Asian languages like Hindi\cite{DBLP:journals/corr/AthavaleBPPS16}, Indonesian\cite{gunawan2018named} and other Indian languages (Kannada, Malayalam, Tamil and Telugu)\cite{gupta2018raiden11}. However, there has been no study on developing neural NER for Nepali language. In this paper, we propose a neural based Nepali NER using latest state-of-the-art architecture based on grapheme-level which doesn't require any hand-crafted features and no data pre-processing. 

%In simple terms, NER task is all about tagging the proper entities (person, location, organization, miscellaneous and others) based on the context of a sentence. 

% We observe that grapheme-level convolutional neural network (CNN) captures better semantic representation compared to character-level CNN in highly inflection language like Nepali.

%Nepali (aka. Khas Kura), is a national language of Nepal and widely spoken by more than 30 million people in Nepal, Assam, Sikkim, Bhutan and Myanmar. Nepali language is written in Devanagari script which dates back to 4th century CE and is direct descendant of Sanskrit language. Nepali language lies under the umbrella of Devanagari language, which consists of other notable languages like Bhojpuri, Bodo, Hindi, Konkani, Marathi, Magahi, Maithili, Sanskrit and many more. These languages are mostly spoken by people in South Asian region.

% lample and hovy did not use POS tag
Recent neural architecture like \cite{DBLP:journals/corr/LampleBSKD16} is used to relax the need to hand-craft the features and need to use part-of-speech tag to determine the category of the entity. However, this architecture have been studied for languages like English, and German and not been applied to languages like Nepali which is a low resource language i.e limited data set to train the model. Traditional methods like Hidden Markov Model (HMM) with rule based approaches\cite{dey2013named},\cite{dey2014named}, and Support Vector Machine (SVM) with manual feature-engineering\cite{bam2014named} have been applied but they perform poor compared to neural. However, there has been no research in Nepali NER using neural network. Therefore, we created the named entity annotated dataset partly with the help of Dataturk\footnote{https://dataturks.com/} to train a neural model. The texts used for this dataset are collected from various daily news sources from Nepal\footnote{https://github.com/sndsabin/Nepali-News-Classifier} around the year 2015-2016.

%NER task on Devanagari scripts is a difficult job because of its inflectional morphology and derivational morphology in nature \cite{bal2004morphological}, \cite{shrestha2016new},\cite{DBLP:journals/corr/AthavaleBPPS16}. Therefore, the hand-crafted rules might not be useful in every domain (business, education, law, medical, official documents) and is very time consuming. Basically, the sentence pattern of English language is Subject + Action Verb + Object, whereas Nepali language pattern is Subject + Object + Action Verb\cite{bal2004structure}. For example, in english, \textbf{Ram goes home.} but in Nepali it is \begin{sanskrit}राम घर जान्छ ।\end{sanskrit}. Here \begin{sanskrit}राम\end{sanskrit} is subject, \begin{sanskrit} घर \end{sanskrit} is object and \begin{sanskrit} जान्छ \end{sanskrit} is action verb.

%There exists  This lack of research in Nepali NER task must be mostly due to unavailability of enough annotated data publicly leading to less motivation among researcher or students.

Following are our contributions:
\begin{enumerate}
    \item We present a novel Named Entity Recognizer (NER) for Nepali language. To best of our knowledge we are the first to propose neural based Nepali NER.
    %Application of state-of-the-art neural architecture model in Nepali language with integration of grapheme-level embedding
    \item As there are not good quality dataset to train NER we release a dataset to support future research
    \item We perform empirical evaluation of our model with state-of-the-art models with relative improvement of upto 10\%
    %Detailed empirical evaluation of different models for Nepali language
    %\item Release of our dataset to boost the NER research in Nepali NLP community
    %\item Achieved 86.7\% F1 score on our dataset without feature engineering or support of gazetteer list.
\end{enumerate}

In this paper, we present works similar to ours in Section \ref{sec:related_work}. We describe our approach and dataset statistics in Section \ref{sec:approach} and \ref{sec:dataset}, followed by our experiments, evaluation and discussion in Section \ref{sec:experiments}, \ref{sec:evaluation}, and \ref{sec:discussion}. We conclude with our observations in Section \ref{sec:conclusion}.

To facilitate further research our code and dataset will be made available at 
github.com/link-yet-to-be-updated
%http://github.com/oya163/oya-nepali-ner.
 
\section{Related Work} \label{sec:related_work}
There has been a handful of research on Nepali NER task based on approaches like Support Vector Machine and gazetteer list\cite{bam2014named} and Hidden Markov Model and gazetteer list\cite{dey2013named},\cite{dey2014named}.

\cite{bam2014named} uses SVM along with features like first word, word length, digit features and gazetteer (person, organization, location, middle name, verb, designation and others). It uses one vs rest classification model to classify each word into different entity classes. However, it does not the take context word into account while training the model. Similarly, \cite{dey2013named} and \cite{dey2014named} uses Hidden Markov Model with n-gram technique for extracting POS-tags. POS-tags with common noun, proper noun or combination of both are combined together, then uses gazetteer list as look-up table to identify the named entities.

%Such traditional sequence labelling algorithms with hand-crafted features and usage of gazetteer lists might not always produce good results on different domains. Also, it is cumbersome to create such domain-specific hand-crafted features and using gazetteer list might not always yield good score as the words might have different tags based on different context. Therefore, there is a need of such algorithm which can have basic understanding of syntactic and semantic knowledge of a human language.

Researchers have shown that the neural networks like CNN\cite{lecun1989backpropagation}, RNN\cite{Rumelhart:1988:LRB:65669.104451}, LSTM\cite{hochreiter1997long}, GRU\cite{DBLP:journals/corr/ChungGCB14} can capture the semantic knowledge of language better with the help of pre-trained embbeddings like word2vec\cite{DBLP:journals/corr/MikolovSCCD13}, glove\cite{pennington-etal-2014-glove} or fasttext\cite{bojanowski2016enriching}. 

Similar approaches has been applied to many South Asian languages like Hindi\cite{DBLP:journals/corr/AthavaleBPPS16}, Indonesian\cite{gunawan2018named}, Bengali\cite{banik2018gru} and 
In this paper, we present the neural network architecture for NER task in Nepali language, which doesn't require any manual feature engineering nor any data pre-processing during training. First we are comparing BiLSTM\cite{hochreiter1997long}, BiLSTM+CNN\cite{DBLP:journals/corr/ChiuN15}, BiLSTM+CRF\cite{DBLP:journals/corr/LampleBSKD16}, BiLSTM+CNN+CRF\cite{DBLP:journals/corr/MaH16} models with CNN model\cite{DBLP:journals/corr/abs-1103-0398} and Stanford CRF model\cite{finkel2005incorporating}. Secondly, we show the comparison between models trained on general word embeddings, word embedding + character-level embedding, word embedding + part-of-speech(POS) one-hot encoding and word embedding + grapheme clustered or sub-word embedding\cite{park-shin-2018-grapheme}. The experiments were performed on the dataset that we created and on the dataset received from ILPRL lab\footnote{http://ilprl.ku.edu.np/}. Our extensive study shows that augmenting word embedding with character or grapheme-level representation and POS one-hot encoding vector yields better results compared to using general word embedding alone.

\section{Approach}\label{sec:approach}
\begin{figure}
    \centering
    \includegraphics[scale=0.6]{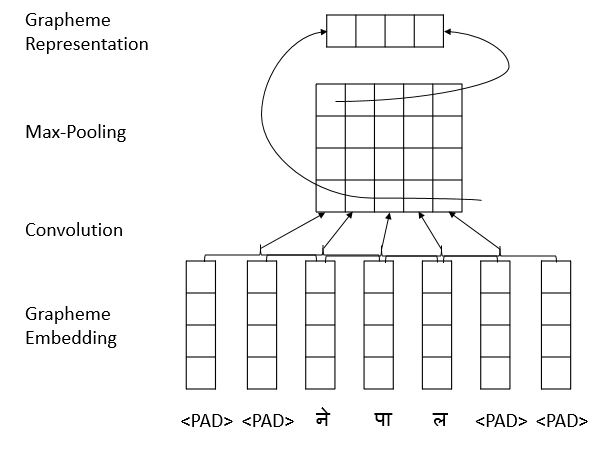}
    \caption{The grapheme level convolution neural network to extract grapheme representation. The dropout layer is applied after maxpooling layer.}
    \label{fig:grapheme}
\end{figure}

\begin{figure*}
    \centering
    \includegraphics[scale=1.0]{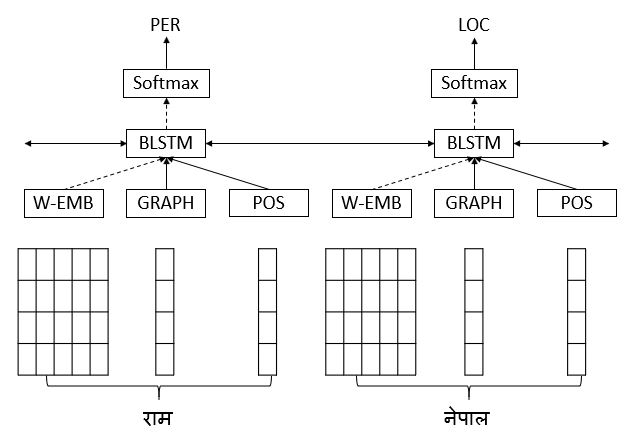}
    \caption{End-to-end model architecture of our neural network. W-EMB, GRAPH, POS represents pre-trained word embeddings, grapheme representations and POS one-hot encoding vectors. GRAPH is obtained from CNN as shown in figure \ref{fig:grapheme}. The dashed line implies the application of dropout.}
    \label{fig:model_architecture}
\end{figure*}

In this section, we describe our approach in building our model. This model is partly inspired from multiple models \cite{DBLP:journals/corr/ChiuN15},\cite{DBLP:journals/corr/LampleBSKD16}, and\cite{DBLP:journals/corr/MaH16}

\subsection{Bidirectional LSTM}
We used Bi-directional LSTM to capture the word representation in forward as well as reverse direction of a sentence. Generally, LSTMs take inputs from left (past) of the sentence and computes the hidden state. However, it is proven beneficial\cite{DBLP:journals/corr/DyerBLMS15} to use bi-directional LSTM, where, hidden states are computed based from right (future) of sentence and both of these hidden states are concatenated to produce the final output as $h_t$=[$\overrightarrow{h_t}$;$\overleftarrow{h_t}$], where $\overrightarrow{h_t}$, $\overleftarrow{h_t}$ = hidden state computed in forward and backward direction respectively.

\subsection{Features}\label{subsec:features}
\subsubsection{Word embeddings}\label{subsubsec:wordemb}
We have used Word2Vec \cite{DBLP:journals/corr/MikolovSCCD13}, GloVe \cite{pennington-etal-2014-glove} and FastText \cite{bojanowski2016enriching} word vectors of 300 dimensions. These vectors were trained on the corpus obtained from Nepali National Corpus\footnote{https://www.sketchengine.eu/nepali-national-corpus/}. This pre-lemmatized corpus consists of 14 million words from books, web-texts and news papers. This corpus was mixed with the texts from the dataset before training CBOW and skip-gram version of word2vec using gensim library\cite{rehurek_lrec}. This trained model consists of vectors for 72782 unique words.

Light pre-processing was performed on the corpus before training it. For example, invalid characters or characters other than Devanagari were removed but punctuation and numbers were not removed. We set the window context at 10 and the rare words whose count is below 5 are dropped. These word embeddings were not frozen during the training session because fine-tuning word embedding help achieve better performance compared to frozen one\cite{DBLP:journals/corr/ChiuN15}.

We have used fasttext embeddings in particular because of its sub-word representation ability, which is very useful in highly inflectional language as shown in Table  \ref{table:embedding}. We have trained the word embedding in such a way that the sub-word size remains between 1 and 4. We particularly chose this size because in Nepali language a single letter can also be a word, for example {\dn e}, {\dn t}, {\dn C}, {\dn r}, {\dn l}, {\dn n}, {\dn u}
and a single character (grapheme) or sub-word can be formed after mixture of dependent vowel signs with consonant letters for example, {\dn {C}} + {\dn {O}} + {\dn {\2}} = {\dn CO\2}, here three different consonant letters form a single sub-word.

\begin{figure*}%
    \centering
    \subfloat[Word2vec embedding]{{\includegraphics[width=7cm]{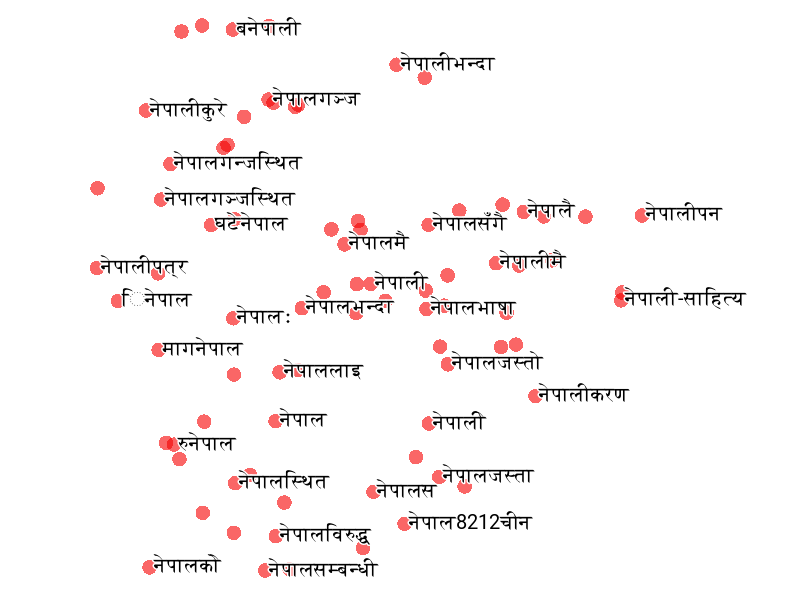} }}%
    \qquad
    \subfloat[Fasttext embedding]{{\includegraphics[width=7cm]{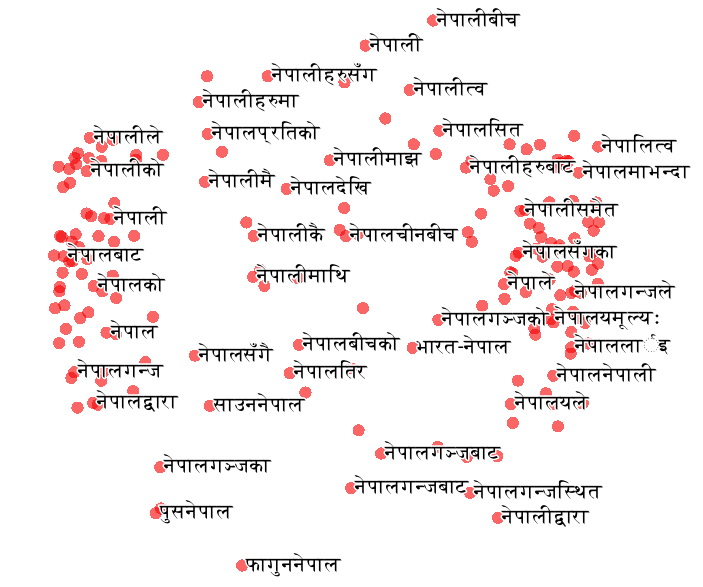} }}%
    \caption{2D Visualization of nearest neighbor word using PCA for a sample word {\dn n\?pAl}}%
    \label{fig:emb_visual}%
\end{figure*}

The two-dimensional visualization of an example word 
{\dn n\?pAl} 
is shown in \ref{fig:emb_visual}. Principal Component Analysis (PCA) technique was used to generate this visualization which helps use to analyze the nearest neighbor words of a given sample word. 84 and 104 nearest neighbors were observed using word2vec and fasttext embedding respectively on the same corpus.

\begin{figure}
    \centering
    \includegraphics[scale=0.6]{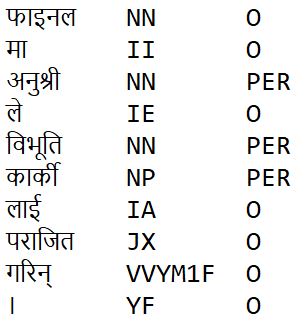}
    \caption{Format of a sample sentence in our dataset.}
    \label{fig:data_format}
\end{figure}

\subsubsection{Character-level embeddings}\label{subsubsec:charemb}
\cite{DBLP:journals/corr/ChiuN15} and \cite{DBLP:journals/corr/MaH16} successfully presented that the character-level embeddings, extracted using CNN, when combined with word embeddings enhances the NER model performance significantly, as it is able to capture morphological features of a word. Figure \ref{fig:grapheme} shows the grapheme-level CNN used in our model, where inputs to CNN are graphemes. Character-level CNN is also built in similar fashion, except the inputs are characters. Grapheme or Character -level embeddings are randomly initialized from [0,1] with real values with uniform distribution of dimension 30.

\subsubsection{Grapheme-level embeddings}\label{subsubsec:graphemb}
Grapheme is atomic meaningful unit in writing system of any languages. Since, Nepali language is highly morphologically inflectional, we compared grapheme-level representation with character-level representation to evaluate its effect. For example, in character-level embedding, each character of a word {\dn n\?pAl} results into {\dn n} + {\dn \?} + {\dn p} + {\dn A} + {\dn l} has its own embedding. However, in grapheme level, a word {\dn n\?pAl} is clustered into graphemes, resulting into {\dn n\?} + {\dn pA} + {\dn l}. Here, each grapheme has its own embedding. This grapheme-level embedding results good scores on par with character-level embedding in highly inflectional languages like Nepali, because graphemes also capture syntactic information similar to characters. We created grapheme clusters using uniseg\footnote{https://uniseg-python.readthedocs.io/en/latest\\/index.html} package which is helpful in unicode text segmentations.
%There are 635 and 447 unique graphemes in \oyeshds dataset and ILPRL\footnote{http://ilprl.ku.edu.np/} dataset respectively.

\subsubsection{Part-of-speech (POS) one hot encoding}\label{subsubsec:posonehot}
We created one-hot encoded vector of POS tags and then concatenated with pre-trained word embeddings before passing it to BiLSTM network. A sample of data is shown in figure \ref{fig:data_format}.

\section{Dataset Statistics}\label{sec:dataset}
\subsection{\oyeshds dataset} \label{sec:nepali-dataset}
Since, we there was no publicly available standard Nepali NER dataset and did not receive any dataset from the previous researchers, we had to create our own dataset. This dataset contains the sentences collected from daily newspaper of the year 2015-2016. This dataset has three major classes Person (PER), Location (LOC) and Organization (ORG). Pre-processing was performed on the text before creation of the dataset, for example all punctuations and numbers besides ',', '-', '|' and '.' were removed. Currently, the dataset is in standard CoNLL-2003 IO format\cite{TjongKimSang:2003:ICS:1119176.1119195}.

Since, this dataset is not lemmatized originally, we lemmatized only the post-positions like {\dn {Ek}}, {\dn {kO}}, {\dn {l\?}}, {\dn {mA}}, {\dn {m\4}}, {\dn {my}}, {\dn {jF}}, {\dn s\1g}, {\dn aEG} which are just the few examples among 299 post positions in Nepali language. We obtained these post-positions from sanjaalcorps\footnote{https://github.com/sanjaalcorps/NepaliStemmer} and added few more to match our dataset. We will be releasing this list in our github repository. We found out that lemmatizing the post-positions boosted the F1 score by almost 10\%.

In order to label our dataset with POS-tags, we first created POS annotated dataset of 6946 sentences and 16225 unique words extracted from POS-tagged Nepali National Corpus and trained a BiLSTM model with 95.14\% accuracy which was used to create POS-tags for our dataset.

The dataset released in our github repository contains each word in newline with space separated POS-tags and Entity-tags. The sentences are separated by empty newline. A sample sentence from the dataset is presented in table \ref{fig:data_format}.

\subsection{ILPRL dataset} \label{sec:ilprl-dataset}
After much time, we received the dataset from Bal Krishna Bal, ILPRL, KU. This dataset follows standard CoNLL-2003 IOB format\cite{TjongKimSang:2003:ICS:1119176.1119195} with POS tags. This dataset is prepared by ILPRL Lab\footnote{http://ilprl.ku.edu.np/}, KU and KEIV Technologies. Few corrections like correcting the NER tags had to be made on the dataset. The statistics of both the dataset is presented in table \ref{table:dataset}.

\begin{table}[!ht]
\centering
\begin{tabular}{|l|l|l|}
\hline
\multicolumn{1}{|c|}{Dataset} & ILPRL & \oyeshds \\ \hline
PER & 262 & 5059 \\ \hline
ORG & 180 & 3811 \\ \hline
LOC & 273 & 2313 \\ \hline
MISC & 461 & 0 \\ \hline
Total entities w/o O & 1176 & 11183 \\ \hline
Others - O & 12683 & 67904 \\ \hline
Total entities w/ O & 13859 & 79087 \\ \hline
Total sentences & 548 & 3606 \\ \hline
\end{tabular}
\caption{Dataset statistics}
\label{table:dataset}
\end{table}

\begin{table}[!ht]
\centering
\begin{tabular}{|l|l|l|}
\hline
\multicolumn{1}{|c|}{Dataset} & ILPRL & \oyeshds \\ \hline
Train & 754 & 7165 \\ \hline
Test & 188 & 2033 \\ \hline
Dev & 234 & 1985 \\ \hline
\end{tabular}
\caption{Dataset division statistics. The number presented are total count of entities token in each set.}
\label{table:dataset_division}
\end{table}

Table \ref{table:dataset_division} presents the total entities (PER, LOC, ORG and MISC) from both of the dataset used in our experiments. The dataset is divided into three parts with 64\%, 16\% and 20\% of the total dataset into training set, development set and test set respectively. 

\begin{table*}[!ht]
\centering
\begin{tabular}{|l|c|c|c|c|c|c|}
\hline
\multirow{3}{*}{Embeddings} & \multicolumn{6}{c|}{\oyeshds} \\ \cline{2-7} 
 & \multicolumn{3}{c|}{Raw} & \multicolumn{3}{c|}{Lemmatized} \\ \cline{2-7} 
 & Train & Test & Val & Train & Test & Val \\ \hline
Random & 78.72 & 63.66 & 64.89 & 88.44 & 75.11 & 77.2 \\ \hline
Word2Vec\_CBOW & 82.33 & 74.59 & 75.15 & 88.05 & 81.96 & 83.82 \\ \hline
Word2Vec\_Skip Gram & 81.58 & 75.93 & 75.75 & 89.84 & 83.47 & 85.79 \\ \hline
GloVe & 87.54 & 76.86 & 76.7 & 92.48 & 82.24 & 84.16 \\ \hline
fastText\_Pretrained & 81.57 & 75.06 & 71.96 & 85.76 & 77.6 & 79.78 \\ \hline
fastText\_CBOW & \multicolumn{1}{l|}{86.01} & \multicolumn{1}{l|}{81.4} & \multicolumn{1}{l|}{80.52} & \multicolumn{1}{l|}{89.23} & \multicolumn{1}{l|}{81.58} & \multicolumn{1}{l|}{83.51} \\ \hline
fastText\_Skip Gram & \multicolumn{1}{l|}{88.31} & \multicolumn{1}{l|}{80.6} & \multicolumn{1}{l|}{78.85} & \multicolumn{1}{l|}{\textbf{94.01}} & \multicolumn{1}{l|}{\textbf{84.74}} & \multicolumn{1}{l|}{\textbf{85.11}} \\ \hline
\end{tabular}
\caption{Effect of different embedding with Bi-LSTM.}
\label{table:embedding}
\end{table*}

\section{Experiments}\label{sec:experiments}
In this section, we present the details about training our neural network. The neural network architecture are implemented using PyTorch framework \cite{paszke2017automatic}. The training is performed on a single Nvidia Tesla P100 SXM2. We first run our experiment on BiLSTM, BiLSTM-CNN, BiLSTM-CRF BiLSTM-CNN-CRF using the hyper-parameters mentioned in Table \ref{table:hyperparameter}. The training and evaluation was done on sentence-level. The RNN variants are initialized randomly from $(-\sqrt{k},\sqrt{k})$ where $k=\frac{1}{hidden\_size}$.

First we loaded our dataset and built vocabulary using torchtext library\footnote{https://torchtext.readthedocs.io/en/latest/}. This eased our process of data loading using its SequenceTaggingDataset class. We trained our model with shuffled training set using Adam optimizer with hyper-parameters mentioned in table \ref{table:hyperparameter}. All our models were trained on single layer of LSTM network. We found out Adam was giving better performance and faster convergence compared to Stochastic Gradient Descent (SGD). We chose those hyper-parameters after many ablation studies. The dropout of 0.5 is applied after LSTM layer.

For CNN, we used 30 different filters of sizes 3, 4 and 5. The embeddings of each character or grapheme involved in a given word, were passed through the pipeline of Convolution, Rectified Linear Unit and Max-Pooling. The resulting vectors were concatenated and applied dropout of 0.5 before passing into linear layer to obtain the embedding size of 30 for the given word. This resulting embedding is concatenated with word embeddings, which is again concatenated with one-hot POS vector.

\subsection{Tagging Scheme}
Currently, for our experiments we trained our model on IO (Inside, Outside) format for both the dataset, hence the dataset does not contain any B-type annotation unlike in BIO (Beginning, Inside, Outside) scheme.

\subsection{Early Stopping}
We used simple early stopping technique where if the validation loss does not decrease after 10 epochs, the training was stopped, else the training will run upto 100 epochs. In our experience, training usually stops around 30-50 epochs.

\subsection{Hyper-parameters Tuning}
We ran our experiment looking for the best hyper-parameters by changing learning rate from (0,1, 0.01, 0.001, 0.0001), weight decay from [$10^{-1}$, $10^{-2}$, $10^{-3}$, $10^{-4}$, $10^{-5}$, $10^{-6}$, $10^{-7}$], batch size from [1, 2, 4, 8, 16, 32, 64, 128], hidden size from [8, 16, 32, 64, 128, 256, 512 1024]. Table \ref{table:hyperparameter} shows all other hyper-parameter used in our experiment for both of the dataset.

\begin{table}[!ht]
\centering
\begin{tabular}{|l|l|}
\hline
\textbf{Hyper-parameters} & \textbf{Value} \\ \hline
LSTM - hidden size        & 100            \\ \hline
CNN - Filter size         & {[}3,4,5{]}    \\ \hline
CNN - Filter number       & 30             \\ \hline
Learning rate             & 0.001          \\ \hline
Weight decay              & 1.00E-006      \\ \hline
Batch size                & 1              \\ \hline
Dropout                   & 0.5            \\ \hline
\end{tabular}
\caption{Hyper-parameters of our experiments}
\label{table:hyperparameter}
\end{table}

\begin{figure}
    \centering
    \includegraphics[scale=0.8]{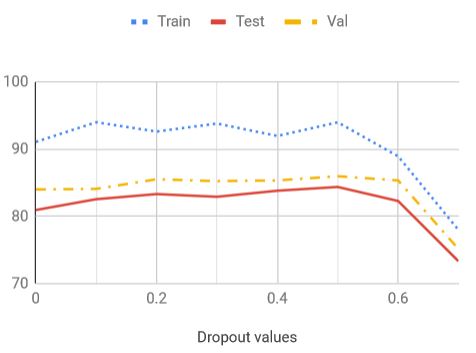}
    \caption{F1 score based on different dropout values using fastText embeddings (Skip Gram). All other hyper-parameter used for this evaluation are presented in table \ref{table:hyperparameter}.}
    \label{fig:dropout}
\end{figure}{}

\subsection{Effect of Dropout}
Figure \ref{fig:dropout} shows how we end up choosing 0.5 as dropout rate. When the dropout layer was not used, the F1 score are at the lowest. As, we slowly increase the dropout rate, the F1 score also gradually increases, however after dropout rate = 0.5, the F1 score starts falling down. Therefore, we have chosen 0.5 as dropout rate for all other experiments performed.

%% DO NOT REMOVE THIS !!!
\iffalse
\begin{table}[ht!]
\centering
\begin{tabular}{|l|c|c|c|}
\hline
 & \multicolumn{3}{c|}{\oyeshds} \\ \hline
Dropout & Train & Test & Val \\ \hline
Not used & 91.1 & 80.97 & 84.03 \\ \hline
0.1 & 94.02 & 82.58 & 84.12 \\ \hline
0.2 & 92.64 & 83.34 & 85.55 \\ \hline
0.3 & 93.83 & 82.94 & 85.27 \\ \hline
0.4 & 91.98 & 83.84 & 85.37 \\ \hline
\textbf{0.5} & \textbf{94.01} & \textbf{84.4} & \textbf{86} \\ \hline
0.6 & 88.96 & 82.31 & 85.37 \\ \hline
0.7 & 78 & 73.35 & 75.17 \\ \hline
\end{tabular}
\caption{F1 score based on different dropout values using fastText embeddings (Skip Gram). All other hyper-parameter used for this evaluation are presented in table \ref{table:hyperparameter} for \oyeshds dataset. {\color{red} Make a line chart with x-axis dropout and y-axis performance.}}
\label{table:dropout}
\end{table}
\fi

\section{Evaluation}\label{sec:evaluation}
In this section, we present the details regarding evaluation and comparison of our models with other baselines.

Table \ref{table:embedding} shows the study of various embeddings and comparison among each other in \oyeshds dataset. Here, raw dataset represents such dataset where post-positions are not lemmatized. We can observe that pre-trained embeddings significantly improves the score compared to randomly initialized embedding. We can deduce that Skip Gram models perform better compared CBOW models for word2vec and fasttext. Here, fastText\_Pretrained represents the embedding readily available in fastText website\footnote{https://fasttext.cc/docs/en/crawl-vectors.html}, while other embeddings are trained on the Nepali National Corpus as mentioned in sub-section \ref{subsubsec:wordemb}. From this table \ref{table:embedding}, we can clearly observe that model using fastText\_Skip Gram embeddings outperforms all other models.

Table \ref{table:model_comparison} shows the model architecture comparison between all the models experimented. 
The features used for Stanford CRF classifier are words, letter n-grams of upto length 6, previous word and next word. This model is trained till the current function value is less than $1\mathrm{e}{-2}$.
The hyper-parameters of neural network experiments are set as shown in table \ref{table:hyperparameter}. Since, word embedding of character-level and grapheme-level is random, their scores are near.

All models are evaluated using CoNLL-2003 evaluation script\cite{TjongKimSang:2003:ICS:1119176.1119195} to calculate entity-wise precision, recall and f1 score.

\begin{figure}
    \centering
    \includegraphics[scale=0.6]{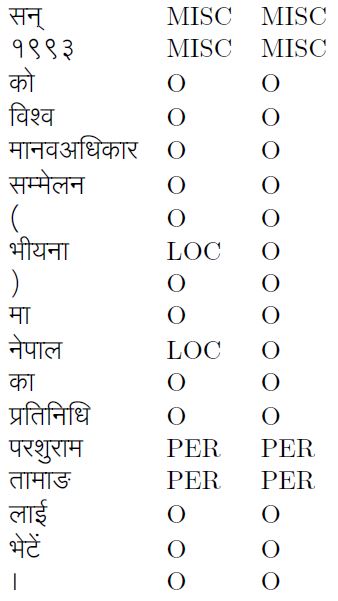}
    \caption{Sample output of the best model from ILPRL test dataset. First, second and third column indicates word to be predicted, ground truth and predicted truth respectively. We can see that not all the tags are classified correctly.}
    \label{fig:sample_output}
\end{figure}

\begin{table*}[ht!]
\centering
\begin{tabular}{|l|c|c|c|c|}
\hline
\textbf{Dataset} & \multicolumn{2}{c|}{\textbf{\oyeshds}} & \multicolumn{2}{c|}{\textbf{ILPRL}} \\ \hline
\textbf{Models} & \textbf{Test} & \textbf{Val} & \textbf{Test} & \textbf{Val} \\ \hline
Stanford CRF & 75.16 & 79.61 & 56.25 & 72.07 \\ \hline
BiLSTM & 84.74 & 85.11 & 80.40 & 83.17 \\ \hline
BiLSTM + POS & 83.65 & 84.09 & 81.25 & 85.39 \\ \hline
BiLSTM + CNN (C) & 86.45 & 87.45 & 80.51 & 85.45 \\ \hline
BiLSTM + CNN (G) & \textbf{86.71} & 86.00 & 78.24 & 83.49 \\ \hline
BiLSTM + CNN (C) + POS & 85.40 & 86.50 & 81.46 & 86.64 \\ \hline
BiLSTM + CNN (G) + POS & 85.46 & 86.43 & \textbf{83.08} & 82.99 \\ \hline

\end{tabular}
\caption{Comparison of different variation of our models}
\label{table:model_comparison}
\end{table*}

\begin{table*}[ht!]
\centering
\begin{tabular}{|l|l|l|l|l|}
\hline
 & \multicolumn{1}{l|}{\textbf{\oyeshds}} & \multicolumn{1}{l|}{\textbf{ILPRL}} \\ \hline
\textbf{Model} & \textbf{Test} & \textbf{Test} \\ \hline
Bam et al. SVM & 66.26 & 46.26 \\ \hline
Ma and Hovy w/ glove & 83.63 & 72.1 \\ \hline
Lample et al. w/ fastText & 85.78 & 82.29 \\ \hline
Lample et al. w/ word2vec & 86.49 & 78.63 \\ \hline
BiLSTM + CNN (G) & \textbf{86.71} & 78.24 \\ \hline
BiLSTM + CNN (G) + POS & 85.46 & \textbf{83.08} \\ \hline
\end{tabular}
\caption{Comparison with previous models based on Test F1 score}
\label{table:previous_model}
\end{table*}

\begin{table*}[ht!]
\centering
\begin{tabular}{|c|c|c|c|c|c|c|}
\hline
\textbf{Dataset} & \multicolumn{3}{c|}{\textbf{\oyeshds}} & \multicolumn{3}{c|}{\textbf{ILPRL}} \\ \hline
\textbf{Entities} & \textbf{Precision} & \textbf{Recall} & \textbf{F1} & \textbf{Precision} & \textbf{Recall} & \textbf{F1} \\ \hline
\textbf{PER} & 93.82 & 88.66 & 91.17 & 74.36 & 72.50 & 73.42 \\ \hline
\textbf{ORG} & 87.28 & 79.59 & 83.26 & 92.31 & 75.00 & 82.76 \\ \hline
\textbf{LOC} & 84.29 & 82.11 & 83.19 & 91.07 & 69.86 & 79.07 \\ \hline
\textbf{MISC} & NA & NA & NA & 94.94 & 87.21 & 90.91 \\ \hline
\textbf{Overall} & 89.45 & 84.14 & \textbf{86.71} & 89.30 & 77.67 & \textbf{83.08} \\ \hline
\end{tabular}
\caption{Entity-wise comparison using best model for respective dataset. MISC-entity is not available for \oyeshds dataset.}
\label{table:entity-wise}
\end{table*}

\section{Discussion}\label{sec:discussion}
In this paper we present that we can exploit the power of neural network to train the model to perform downstream NLP tasks like Named Entity Recognition even in Nepali language. We showed that the word vectors learned through fasttext skip gram model performs better than other word embedding because of its capability to represent sub-word and this is particularly important to capture morphological structure of words and sentences in highly inflectional like Nepali. This concept can come handy in other Devanagari languages as well because the written scripts have similar syntactical structure.

We also found out that stemming post-positions can help a lot in improving model performance because of inflectional characteristics of Nepali language. So when we separate out its inflections or morphemes, we can minimize the variations of same word which gives its root word a stronger word vector representations compared to its inflected versions.

We can clearly imply from tables \ref{table:dataset}, \ref{table:dataset_division}, and \ref{table:model_comparison} that we need more data to get better results because \oyeshds dataset volume is almost ten times bigger compared to ILPRL dataset in terms of entities.

\section{Conclusion and Future work}\label{sec:conclusion}
In this paper, we proposed a novel NER for Nepali language and achieved relative improvement of upto 10\% and studies different factors effecting the performance of the NER for Nepali language. 

%we show that we can use deep learning techniques to improve the sequence labelling tasks in Nepali language with out any handcrafted features or external support of gazetteers list. We present the state-of-the-art performance with 83.91 F1 score in named entity recognition tasks in Nepali language even under data-scarce situation.

We also present a neural architecture BiLSTM+CNN(grapheme-level) which turns out to be performing on par with BiLSTM+CNN(character-level) under the same configuration. We believe this will not only help Nepali language but also other languages falling under the umbrellas of Devanagari languages. Our model BiLSTM+CNN(grapheme-level) and BiLSTM+CNN(G)+POS outperforms all other model experimented in \oyeshds and ILPRL dataset respectively.

Since this is the first named entity recognition research in Nepal language using neural network, there are many rooms for improvement. We believe initializing the grapheme-level embedding with fasttext embeddings might help boosting the performance, rather than randomly initializing it. In future, we plan to apply other latest techniques like BERT, ELMo and FLAIR to study its effect on low-resource language like Nepali. We also plan to improve the model using cross-lingual or multi-lingual parameter sharing techniques by jointly training with other Devanagari languages like Hindi and Bengali.

Finally, we would like to contribute our dataset to Nepali NLP community to move forward the research going on in language understanding domain. We believe there should be special committee to create and maintain such dataset for Nepali NLP and organize various competitions which would elevate the NLP research in Nepal.

Some of the future works are listed below:
\begin{enumerate}
    \item Proper initialization of grapheme level embedding from fasttext embeddings.
    \item Apply robust POS-tagger for Nepali dataset
    \item Lemmatize the \oyeshds dataset with robust and efficient lemmatizer
    \item Improve Nepali language score with cross-lingual learning techniques
    \item Create more dataset using Wikipedia/Wikidata framework
\end{enumerate}

\section{Acknowledgments}
The authors of this paper would like to express sincere thanks to Bal Krishna Bal, Kathmandu University Professor for providing us the POS-tagged Nepali NER data.

\bibliography{emnlp-ijcnlp-2019}
\bibliographystyle{acl_natbib}

\end{document}